\documentclass[10pt,twocolumn,letterpaper]{article}

\usepackage{cvpr}
\usepackage{times}
\usepackage{epsfig}
\usepackage{graphicx}
\usepackage{amsmath}
\usepackage{amssymb}
\usepackage{subfigure}
\usepackage{algorithm}
\usepackage{algpseudocode}
\usepackage{multirow}
\usepackage{commath}
\usepackage{booktabs}
\newcommand\blfootnote[1]{%
  \begingroup
  \renewcommand\thefootnote{}\footnote{#1}%
  \addtocounter{footnote}{-1}%
  \endgroup
}

\usepackage[pagebackref=true,breaklinks=true,colorlinks,bookmarks=false]{hyperref}

\cvprfinalcopy 

\def\cvprPaperID{4635} 

\ifcvprfinal\fi
\begin{document}

\title{FOAL: Fast Online Adaptive Learning for Cardiac Motion Estimation}

\author{Hanchao Yu$^{\star}$\textsuperscript{2},
Shanhui Sun$^{\dagger}$\textsuperscript{1},
Haichao Yu\textsuperscript{2}, 
Xiao Chen\textsuperscript{1},\\
Honghui Shi$^{\dagger}$\textsuperscript{3}, 
Thomas Huang\textsuperscript{2},
Terrence Chen\textsuperscript{1}
\\
{\normalsize $^{1}$United Imaging Intelligence, Cambridge, MA 02140},
{\normalsize $^{2}$University of Illinois at Urbana-Champaign},
{\normalsize $^{3}$University of Oregon} \\
{\normalsize $^{\dagger}$~{\tt shanhui.sun@united-imaging.com}, $^{\dagger}$~{\tt shihonghui3@gmail.com}}
}

\maketitle
\thispagestyle{empty}
\blfootnote{$^{\star}$ This work was carried out during the internship of the author at United Imaging Intelligence, Cambridge, MA 02140.}
\begin{abstract}
Motion estimation of cardiac MRI videos is crucial for the evaluation of human heart anatomy and function. Recent researches show promising results with deep learning-based methods. In clinical deployment, however, they suffer dramatic performance drops due to mismatched distributions between training and testing datasets, commonly encountered in the clinical environment. On the other hand, it is arguably impossible to collect all representative datasets and to train a universal tracker before deployment. In this context, we proposed a novel fast online adaptive learning (FOAL) framework: an online gradient descent based optimizer that is optimized by a meta-learner. The meta-learner enables the online optimizer to perform a fast and robust adaptation. We evaluated our method through extensive experiments on two public clinical datasets. The results showed the superior performance of FOAL in accuracy compared to the offline-trained tracking method. On average, the FOAL took only $0.4$ second per video for online optimization.  
\end{abstract}

\vspace{-5mm}
\section{Introduction}

Video dense tracking and motion estimation using deep learning has gained great progress for natural image applications in recent research~\cite{sun2018pwc, ilg2017flownet, meister2018unflow, yin2018geonet, zhai2019optical, Lai_2019_CVPR, Hur_2019_CVPR, Wang_2019_CVPR, Lei_2019_CVPR, Ranjan_2019_CVPR, Zhu_2019_CVPR, Liu_2019_CVPR}. In medical imaging, videos,  compared to static images, are ideal for dynamically changing physiological processes such as the beating heart and are commonly used in clinical settings. Feature tracking of dynamic cardiac images can provide precise and comprehensive assessments of the cardiac motion and has been proved valuable for cardiac disease management~\cite{spath2019global, reindl2019prognostic, xia2019tracking, padiyath2013echocardiography}. Motion estimation can also benefit other tasks in cardiac imaging, such as image reconstruction~\cite{huang2019dynamic, seegoolam2019exploiting} and semi-supervised segmentation~\cite{qin2018joint, valvano2019temporal, zheng2019explainable, lee2019tetris, yang2019deep,wei2018revisiting}. Recently, deep learning-based methods show promising results in cardiac motion estimation~\cite{qin2018joint, zheng2019explainable, krebs2019learning, morales2019implementation}. However, most studies have been designed in a research environment: the proposed models are trained and tested on the data with similar distributions. In a clinical environment, however, the imaged objects may present various anatomies (abnormally thin or thick heart muscle) and/or dynamics (irregularly beating heart) for different diseases. On top of that, the imaging process itself commonly introduces many, if not more, variations. This is especially true for cardiac magnetic resonance (CMR) imaging, which provides superior video quality over ultrasound, but the image appearances are influenced by multiple factors including scanner vendors, main magnetic fields, different scanning protocols and technicians' operations. It is arguably impossible to build a dataset that includes every combination of the variations and train a universal tracker on it. It is also not ideal and sometimes impossible in a clinical setting that the pre-trained network gets fine-tuned on the data from a different distribution, given the scarce nature of medical data. In other words, for a clinically suitable deep-trained tracker, the neural network needs to possess the capability to quickly adapt to new data from unseen distributions. Towards this end, we propose a fast online adaptive learning (FOAL) mechanism for dense video tracking applied to cardiac motion estimation. The proposed framework consists of an online adaptive stage and an offline meta-learning stage. The offline meta-learning trains the model to gain the adaptation capability and the online stage will apply this adaptation to adjust the model parameters using very few and unseen data. We have designed a unique module for video tracking used in both stages to train an adaptive tracker. The tracker trained using the proposed FOAL achieves the state-of-the-art (SOTA) results compared to strong baselines. The contributions of our work are summarized as follows.
\begin{itemize}
  \item In the context of dense motion estimation, we proposed a novel online model adaptation method, which adapts a trained baseline model to a new video using a gradient descent optimization. 
  \item We proposed a meta-learning method optimizing the proposed online optimizer. The meta-learner enables the online optimizer to perform a fast and robust adaption.
  \item We proposed practical solutions for training meta learner in dense motion estimation task.
  \item Our proposed method is not limited to the network structure of the baseline dense motion estimation. The extensive experiments consistently demonstrated superior performance improvement of our method in accuracy comparing to the baseline model. 
\end{itemize}
\begin{figure}[ht]
\begin{center}
  \includegraphics[width=0.96\linewidth]{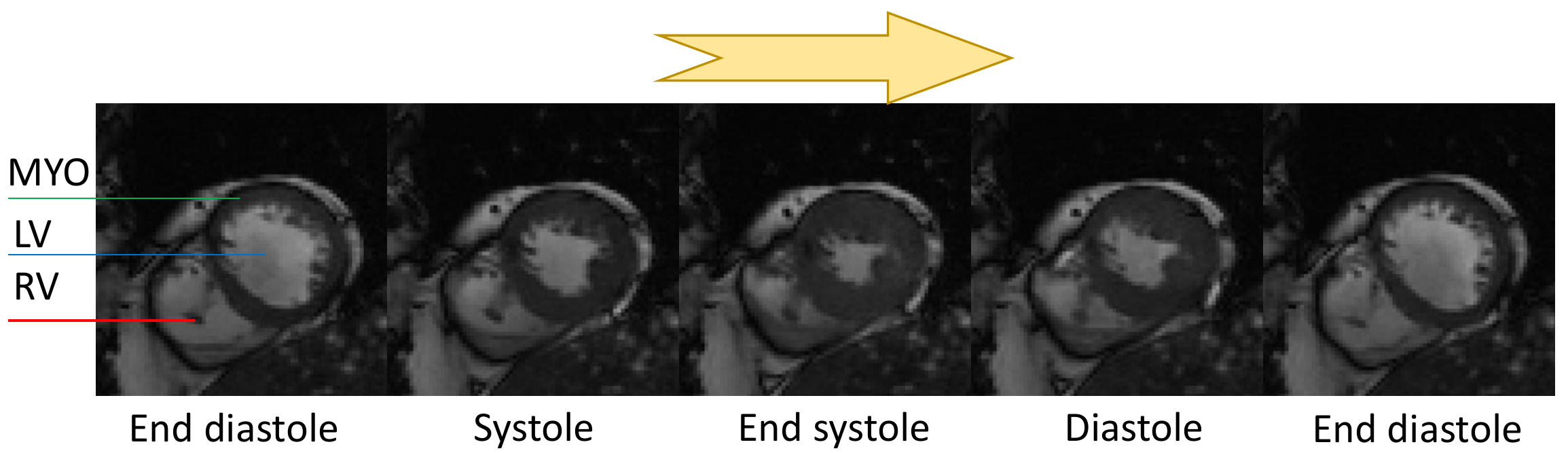}
\end{center}
  \caption{A typical cardiac cycle of a healthy subject recorded by CMR. The cycle indicates the heart relaxation and contraction process. The myocardium (MYO) appears as a dark ring in the image. The left ventricle (LV) is filled with a hyperintense blood signal contained inside the ring. The right ventricle (RV) cavity is indicated via the red line.}
\label{heart cycle}
\end{figure}
\vspace{-5mm}
\section{Related Work}
Section~\ref{related:natural} discusses state-of-the-arts in the literature for motion estimation in the computer vision field. Section~\ref{related:cardiac} introduces the task of cardiac motion estimation and existing studies on this topic. Section~\ref{related:maml} introduces the model-agnostic meta-learning which has inspired our method. 

\subsection{Motion Estimation for Camera Videos}\label{related:natural}
Motion estimation is one of the fundamental problems in the computer vision field. In the literature, there are a few deep learning-based approaches solving motion estimation such as reported works in ~\cite{cheng2017segflow, dosovitskiy2015flownet, ilg2017flownet, sun2018pwc, meister2018unflow}. Dosovitskiy~\etal~\cite{dosovitskiy2015flownet} proposed two optical flow estimation networks (Flownets): FlownetSimple and FlownetCorr. The former is a generic architecture and the latter includes a correlation layer to fuse feature vectors at different image locations. Flownet 2.0 in the work \cite{ilg2017flownet} further adds an extra branch to deal with pairs with small displacement and uses the original Flownet to deal with large displacement. Sun \etal~\cite{sun2018pwc} proposed a smaller and more efficient neural network structure utilizing feature pyramid as well as cost volume to get a more accurate motion. Most of these above works used a supervised learning approach with true motion fields. In contrast to these supervised methods, Meister~\etal~\cite{meister2018unflow} proposed an unsupervised framework where the flow was predicted and used to warp the source image to the reference image. The model is optimized to minimize the difference between the warped image and the reference image. Besides, an occlusion-aware forward-backward consistency loss is used with the census transform to improve the tracking results. Note that our baseline model utilized a similar self-supervision idea as ~\cite{meister2018unflow}.  

\subsection{Cardiac Motion Estimation}\label{related:cardiac}

Cardiac motion estimation takes a time series (video) of CMR images as input and predicts the heart motion through time. Motion fields are usually estimated at a pixel level due to the non-rigid nature of cardiac contraction. Normally the video records a complete cardiac contraction cycle: from the onset of contraction (end-diastolic ED), then to maximum contraction (end-systolic ES) and back to relaxation. Fig.~\ref{heart cycle} shows example CMR frames from a video of a normal subject. The motion of a frame is usually estimated relative to a reference frame that is commonly chosen as the ED or ES frame. Let frame at time $t$ be $I(x,y,t)$, and $I(x,y,t_{ref})$ as the reference image. The goal of motion estimation is to find the mapping $F_{\theta}$ such that 
\begin{equation}
    F_{\theta}: (I(x, y, t_{ref}), I(x, y, t)) \longrightarrow  V_x(x, y, t), V_y(x, y, t)
    \label{formulation}
\end{equation}
where $F_{\theta}$ is the mapping function with parameter $\theta$ and $V_x, V_y$ are the motion fields along $x$ and $y$ directions, respectively. Motion tracking methods can be generally categorized according to different formulations of $F_{\theta}$: optical flow based, conventional image registration based, and deep learning based. 

\begin{figure}[ht]
\centering
\subfigure[]{
\label{bright} 
\includegraphics[width=0.2\linewidth]{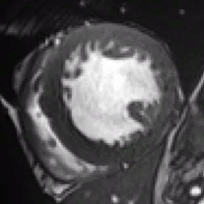}}
\subfigure[]{
\label{dark} 
\includegraphics[width=0.2\linewidth]{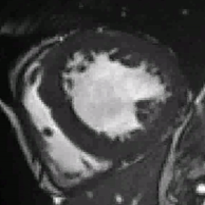}}
\subfigure[]{
\label{motion1} 
\includegraphics[width=0.2\linewidth]{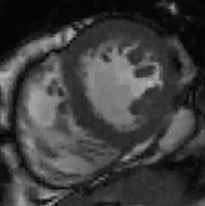}}
\subfigure[]{
\label{motion2} 
\includegraphics[width=0.2\linewidth]{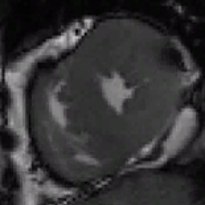}}
\caption{Examples of challenges in CMR motion estimation. (a) and (b) are from one CMR video, where the upper part of the LV myocardium (anterior wall) has a big intensity drop due to the changes in MR coil detection sensitivity. (c) and (d) are from another CMR video, where large motion occurs between an end-diastolic frame (source) and an end-systolic frame (reference).}
\label{violation} 
\end{figure}

The optical flow based method is built on several presumptions on image appearance and motion strength, such as brightness consistency and small motion between the source and reference frames. The problem of applying optical flow based methods to CMR motion estimation is that the presumptions are violated in CMR videos~\cite{earls2002cardiac}. Fig.~\ref{violation} shows some example images, illustrating the challenges of CMR. Wang~\etal~\cite{wang2019gradient} proposed a novel gradient-flow based method that uses a local shape model to keep the local intensity and shape features invariance. 

In addition to the optical flow based approaches, image registration based methods~\cite{puyol2018fully, rueckert1999nonrigid, de2012temporal, shen2005consistent, shi2012comprehensive, tobon2013benchmarking, krebs2019learning,yu2019novel} were applied to solve cardiac motion estimation. Craene \etal~\cite{de2012temporal} utilized B-spline velocity fields with physical constraints to compute the trajectories of feature points and performed the tracking. Rueckert \etal~\cite{rueckert1999nonrigid} proposed a free form deformation (FFD) method solving a general deformable image registration problem and recent work~\cite{puyol2018fully, shen2005consistent, shi2012comprehensive, tobon2013benchmarking} utilize this method to estimate the cardiac motion. It is known that FFD-like methods suffer from the computation efficiency problem. To address this issue, Vigneault \etal~\cite{vigneault2017feature} proposed a coarse-to-fine registration framework to track cardiac boundary points. This solution improved the time efficiency but an extra segmentation step was required. In addition, this sparse tracking lost motion understanding in the heart muscle region. 

Recent success in deep neural network solving many computer vision problems has inspired efforts to explore deep learning based cardiac motion estimation. Qin~\etal~\cite{qin2018joint} proposed a multi-task framework that combines segmentation and motion estimation tasks. The learned cardiac motion field is used to warp the segmentation mask and guide the segmentation module in a semi-supervised manner. The results show that both segmentation and motion estimation performance is improved compared to a single task. Zheng~\etal~\cite{zheng2019explainable} proposed the apparent flow net which is a modified U-net. The segmentation masks were used in the apparent flow net to improve motion estimation. In work~\cite{krebs2019learning}, a conditional variational autoencoder (VAE) based method was presented to estimate the cardiac motion. The VAE encoder is used to map deformations to latent variables, which is regularized via Gaussian distribution and decode to a deformation filed via VAE decoder. Note that it is generally hard to obtain true cardiac motion and thus above works were quantitatively evaluated using the segmentation masks. In this work, we also use this type of evaluation. 

\subsection{Model Agnostic Meta Learning}\label{related:maml}
Meta-learning, or learning to learn, aims to build a universal meta-model that could make fast adaptation to new tasks~\cite{schmidhuber1987evolutionary}. Model-agnostic meta-learning (MAML)~\cite{finn2017model} is a general strategy that searches for good model-agnostic initialization parameters that are trained through training tasks and can quickly adapt to new tasks. Given the initial model parameters $\theta$, for every task $T_i$ in the training set, the task-specific parameters $\theta_i$ are independently updated within the task dataset using gradient descent with a differentiable loss function $L$:
\begin{equation}
    \theta_i \leftarrow \theta - \alpha \nabla_{\theta}L(T_i;\theta).
\end{equation}

Then the original model parameters $\theta$ are updated over all the training tasks:
\begin{equation}
    \theta \leftarrow \theta - \beta  \nabla_{\theta}\sum_{i}L(T_i;\theta_i).
\end{equation}

Through these meta-training processes, the optimal ``initialization" parameters are supposed to be sensitive to new task adaptation within a limited number of adaptation steps. MAML has been widely used in few-shot learning~\cite{finn2017one, sun2019learning, garcia2017few}, neural architecture search~\cite{liu2018darts}, graphical neural network~\cite{garcia2017few}, compressed sensing~\cite{wu2019deep} and transfer learning~\cite{zamir2018taskonomy}. Most applications using MAML are to solve high-level vision tasks such as classification and recognition. The MAML method inspired us to utilize a meta learner which teaches the model to learn how to adapt to a new video. 

\section{Method}
We proposed an online adaptive tracking framework in the context of dense motion tracking utilizing a deep neural network. The proposed method is a general video tracking framework that is not limited to motion estimation in CMR. Nevertheless, without loss of generality, the method is presented in the CMR context.

\begin{figure}[ht]
\begin{center}
  \includegraphics[width=0.95\linewidth]{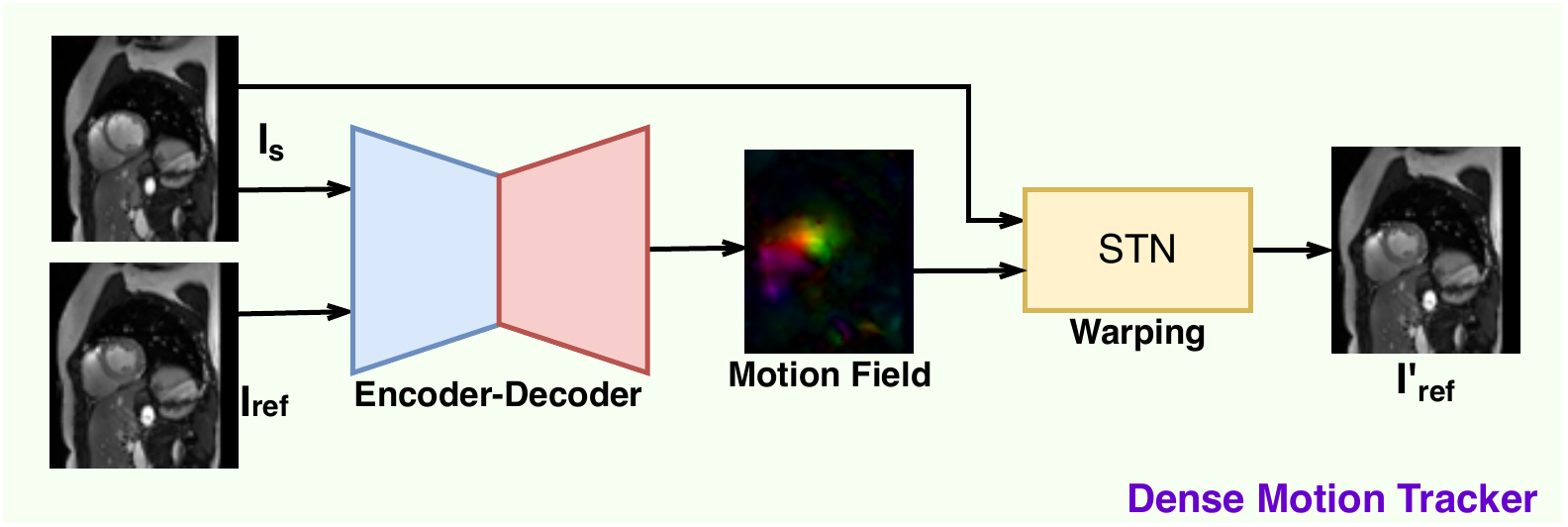}
\end{center}
  \caption{Overview of the dense tracking framework. The encoder is a Siamese structure that takes source and reference images as input. The feature maps produced by the Siamese encoder are concatenated and fed into the decoder. }
\label{fig:backbone}
\end{figure}
\vspace{-3mm}
\subsection{Dense Motion Tracking}\label{sec:dense_tracker}
Fig.~\ref{fig:backbone} depicts the architecture of our dense tracking framework. The overall idea of the dense motion tracking is an end-to-end unsupervised learning approach that inspired from~\cite{meister2018unflow}. Annotating the motion field for the heart is an intractable task and unsupervised learning avoids the necessity of the ground truth. In our work we used a lightweight backbone of the network: the inputs are source image and reference image (e.g. two frames in the same video). The encoder is a Siamese~\cite{chopra2005learning} structure. The decoder is a series of convolution and transpose convolution operators used to decode the features and restore the output to the original image size. The output is the predicted motion field. To perform unsupervised learning, the spatial transformer network~\cite{jaderberg2015spatial} is utilized to deform/warp the source image to the reference image and image reconstruction loss $L_{mse}$ is used to minimize the difference between the warped source image and the reference image. $L_{mse}$ is the mean square error (MSE). In addition to $L_{mse}$, motion field smoothness $L_{smooth}$ proposed in~\cite{qin2018joint} is used to avoid abrupt motion change and a bidirectional (forward-backward) flow consistency loss $L_{con}$ proposed in~\cite{meister2018unflow} is used. The total loss $L_{total}$ is thus defined as follows:
\begin{equation}
    L_{total} = L_{mse} + \alpha_{s} L_{smooth} + \beta_{c} L_{con},
    \label{eq:total_loss}
\end{equation}
where $\alpha_{s}$ and $\beta_{c}$ are used to balance the three losses. 


\subsection{Online Optimizer}
The unsupervised dense tracking (Section~\ref{sec:dense_tracker}) mitigates the need for ground truth motion fields. However, the distribution mismatch between training and test datasets is a continuous challenge, particularly the long tail problem in the medical image domain. The clinical deployment of a deep learning model suffers the domain mismatch problem. It is a challenge to collect sufficient samples to train a universal tracker. In this section, in the context of the proposed dense tracking, we extend the tracker to address the dataset distribution mismatch problem. Instead of training such a universal tracker offline, we make the tracker being aware of the test data online. The idea behind this is to enable a given tracker to automatically adapt to a new video $x$. Suppose we have a model $f_{\theta}$ using the proposed dense tracker trained on dataset $D_{a}$ with a distribution $p(D_{a})$. The online adaptive learning on video $x$ is an online optimization algorithm and is realized via back-propagating through the stochastic gradient descent steps as follows:

\begin{equation}
\theta' \leftarrow \theta' -\alpha\nabla_{\theta'}L(f_{\theta'}),
\label{eq:adaptive_learning_step}
\end{equation}
where $\theta'$ represents the model parameters and is initialized from $\theta$. $\alpha$ is the learning rate. We utilized the same loss function $L$ defined in Eq.\eqref{eq:total_loss}. The overview of the online adaptive algorithm is outlined in the Algorithm~\ref{alg:online}.
\begin{algorithm}[ht]
\caption{FOAL online optimization}
\label{alg:online}
\begin{algorithmic}
\Require Single video $k$: $x_k$, learning rate: $\alpha$, trained model: $f_{\theta}$, number of online tracking optimization steps: $M$
    \State $\theta'_k \leftarrow \theta$
    \State Sample $K$ pairs $D_{k}=\{a_k^{(j)}, b_k^{(j)}\}$ from video $x_k$
    \For {$m$ from $1$ to $M$}
        \State Evaluate loss $L_m(f_{\theta'_k})$ using $D_{k}$
        \State Compute parameters with gradient descent:
        \State $\theta'_k \leftarrow \theta'_k -\alpha\nabla_{\theta'_k}L_m(f_{\theta'_k})$
    \EndFor
\Ensure updated network weights: $\theta'_k$
\end{algorithmic}
\end{algorithm}
\vspace{-2mm}

It is worth pointing out that the gradient descent steps are performed over all parameters of the network at the online stage. Thus, it is computationally expensive to optimize them on all image pairs (source and reference) with too many steps. We aim to adapt the offline model in just a few steps using only a small number of online samples. We realize this by utilizing meta-learning to optimize this optimization procedure. This idea is inspired by MAML~\cite{finn2017model}, which is used to learn good initial model parameters via meta-learning. Like in MAML, we perform a second-order optimization by back-propagation using stochastic gradient descent through the online optimization Eq.~\eqref{eq:adaptive_learning_step}. 

\begin{algorithm}[tb]
\label{alg:meta_train}
\caption{FOAL offline meta-learning}
\label{alg:meta-train}
\begin{algorithmic}
\Require video set: $X$, learning rate: $\alpha$, $\beta$, initial model: $f_{\theta}$, number of online tracking optimization steps: $m$
\While {not done}
\State Sample $N$ videos $\{x_1,x_2,...,x_N\}$ from $X$
\For{$i$ from $1$ to $N$}
    \State $\theta'_i \leftarrow \theta$
    \State Sample $K$ pairs $D_{i}=\{a_i^{(j)}, b_i^{(j)}\}$ from video $x_i$
    \For {$t$ from $1$ to $m$}
        \State Evaluate loss $L_i(f_{\theta'_i})$ using $D_{i}$
        \State Compute parameters with gradient descent:
        \State $\theta'_i \leftarrow \theta'_i -\alpha\nabla_{\theta'_i}L_i(f_{\theta'_i})$
    \EndFor
    \State Sample $K$ pairs $D'_{i}=\{a_i^{(k)}, b_i^{(k)}\}$ from video $x_i$
\EndFor
\State Model update: $\theta \leftarrow \theta -\beta \nabla_{\theta}\frac{1}{N} \sum_{i}^{N}L_i(f_{\theta'_i})$ using each $D'_{i}$ and video-specific loss $L_i(f_{\theta'_i})$
\EndWhile
\Ensure updated model $\theta$
\end{algorithmic}
\end{algorithm}

\subsection{Meta-learning}
We utilized a meta leaner to re-train the model $f_{\theta}$ on the dataset $D_{meta}$ from parameters $\theta$ in order to teach the online optimizer in Eq.~\eqref{eq:adaptive_learning_step}. The optimizer learns to adapt $f_{\theta}$ for a given video $x$. Note that $D_{meta}$ is either $p(D_a)$ or a new distribution $p(D_b)$, where $D_b$ is a new dataset, and $p(D_b)$ may mismatch domain $p(D_a)$. The full algorithm is outlined in Algorithm~\ref{alg:meta-train}. There are two For-loops in Algorithm~\ref{alg:meta-train}. The inner For-loop is the proposed optimization algorithm in Algorithm~\ref{alg:online} to optimize the online optimizer Eq.~\eqref{eq:adaptive_learning_step}. The outer For-loop is the meta-leaner and the meta optimizer is defined as follows.

\begin{equation}
    \theta \leftarrow \theta -\beta \nabla_{\theta}\frac{1}{N}\sum_{i}^{N}L_i(f_{\theta'_i}),
    \label{eq:meta_opt}
\end{equation}
where $i$ is $i^{th}$ video in the training procedure. $N$ is the number of videos in a batch size for optimizing the meta learner. $\beta$ is the learning rate of the meta-learner. $L_i$ is the loss (Eq.~\ref{eq:total_loss}) evaluated on the $i^{th}$ video. $f_{\theta'_i}$ is the model parameters for the $i^{th}$ video.

\subsection{Practical Version of the Meta-Learning}
\textbf{Memory limitation and solution:} In contrast to few-shot learning (a classification problem) discussed in MAML~\cite{finn2017model}, dense motion tracker need store a larger number of feature maps (i.e. requiring a large amount of GPU memory) given a larger image size (e.g. $192 \times 192$). The meta optimizer (Eq.~\ref{eq:meta_opt}) requires computing derivatives of each independent model associated with a specific video. To tackle this problem, by employing the property that the gradient operator and the average operator are commutative in Eq.~\ref{eq:meta_opt}, we swap the two operators as shown in Eq.~\eqref{eq:trick_memory}. %
\begin{equation}
    \nabla_{\theta}\frac{1}{N}\sum_{i}^{N}L_i(f_{\theta'_i}) \Leftrightarrow \frac{1}{N}\sum_{i}^{N}\nabla_{\theta}L_i(f_{\theta'_i})
    \label{eq:trick_memory}
\end{equation}
which enables computing gradients on GPU and transferring them to CPU. 

\textbf{First order derivative approximation:} Note that in Eq.~\eqref{eq:trick_memory}, second-order derivative is needed in back-propagation. This involves calculating the second-order Hessian matrix, which is computationally costly. As a workaround, we use first-order approximation, whose effectiveness is demonstrated in MAML~\cite{finn2017model}. In~\cite{finn2017model}, the approximation rendered comparable results to the second-order derivatives.

\begin{table}[tb]
\small
 \caption{Inside distribution v.s. outside distribution Dice coefficient results for the baseline model, proposed FOAL without meta-learning (FOAL w/o meta) and proposed FOAL with meta-learning (FOAL + meta). Averaged Dice coefficient with standard deviation is given among five-fold leave-one-disease-out cross-validation.}
 \label{tab:acdc_dice}
 \centering
 \setlength{\tabcolsep}{1.7mm}
 \begin{tabular}{cccc}
  \toprule
  \multirow{2}{*}{Method}   & LV& RV & MYO \\
  \cmidrule(r){2-4} & \multicolumn{3}{c}{Inside Distribution Test Set}\\
  \midrule
 Baseline &0.838(0.024) & 0.825(0.013) & 0.797(0.014) \\
 FOAL w/o meta &   0.856(0.021) & 0.842(0.013) & {0.820(0.008)}\\
 FOAL + meta & \textbf{0.873(0.019)} & \textbf{0.859(0.013)} & \textbf{0.840(0.007)}  \\
 \hline
      & \multicolumn{3}{c}{Outside Distribution Test Set}\\
 \hline 
Baseline  & 0.840(0.094) & 0.775(0.096) & 0.803(0.045) \\
 FOAL w/o meta & 0.863(0.077) & 0.801(0.085) & 0.828(0.031)\\
 FOAL + meta & \textbf{0.880(0.065)} & \textbf{0.806(0.086)} & \textbf{0.846(0.027)} \\      
  \bottomrule
 \end{tabular}
\end{table}


\begin{table}[tb]
\small
 \caption{Inside distribution v.s. outside distribution Hausdorff distance (mm) results for the baseline model, proposed FOAL without meta-learning (FOAL w/o meta) and proposed FOAL with meta-learning (FOAL + meta). Averaged Hausdorff distance with standard deviation is given among five-fold leave-one-disease-out cross-validation.}
 \label{tab:acdc_hd}
 \centering
 \setlength{\tabcolsep}{1.5mm}
 \begin{tabular}{cccc}
  \toprule
  \multirow{2}{*}{Method}   & LV& RV & MYO \\
  \cmidrule(r){2-4} & \multicolumn{3}{c}{Inside Distribution Test Set}\\
  \midrule
 Baseline & 7.265(0.779) & 8.782(0.422) & 6.930(0.548) \\
 FOAL w/o meta &   6.417(0.627) & 8.141(0.329) & {6.286(0.469)}\\
 FOAL + meta & \textbf{6.012(0.580)} & \textbf{7.731(0.303)} & \textbf{6.157(0.489)} \\
 \hline
      & \multicolumn{3}{c}{Outside Distribution Test Set}\\
 \hline 
Baseline & 6.921(2.147) & 10.173(1.436) & 6.716(1.803) \\
FOAL w/o meta & 6.158(1.727) & 9.320(1.422) & 6.107(1.506)\\
 FOAL + meta & \textbf{5.832(1.534)} & \textbf{9.378(1.417)} & \textbf{5.987(1.437)} \\      
  \bottomrule
 \end{tabular}
\end{table}

\section{Evaluation Methodology}
In this section, we present evaluation methodology on compared tracking methods: tracking performed using proposed dense motion tracking method (baseline model), tracking performed using online optimization from the baseline model without meta-learning (FOAL without meta-learning), and tracking performed using online optimization with meta-learning (FOAL with meta-learning). 
\vspace{-3mm}
\subsection{Datasets and Evaluation Reference}
In our study, two public CMR datasets were utilized: ACDC dataset~\cite{bernard2018deep} and Kaggle Data Science Bowl Cardiac Challenge Data~\cite{kaggle}. All data acquisitions were performed using breath-holding so that only cardiac motion is observed in the videos. It is arguably impossible to make an independent reference standard of the cardiac motion manually. To perform quantitative analysis, we utilized segmentation masks as the independent reference standard. In the test dataset of the study, we have heart segmentation references at both the first frame and the evaluated reference frame. We generate the segmentation masks via warping source segmentation to the reference and compare it to the annotation using quantitative indices defined in section~\ref{sec:quant_indices}.

\vspace{0.2em}\noindent\textbf{ACDC Dataset:}\hspace{0.5em} It includes short-axis view CMR videos from 100 subjects (healthy and diseased cases). Each subject contains multiple slices (9-10) and each slice is a video sequence covering at least one heartbeat cycle. Overall, there are 951 videos in this dataset. Each video provides two heart segmentation masks: one for the ED phase and one for the ES phase. The segmentation labels are right ventricle (RV) cavity, myocardium (MYO) and left ventricle (LV) cavity. In addition, the 100 subjects are evenly divided into 5 categories with 20 subjects each. These are diagnosed into: normal cases (NOR),  systolic heart failure with infarction (MINF),  dilated cardiomyopathy (DCM), hypertrophic cardiomyopathy (HCM), abnormal right ventricle (ARV). The CMR videos were collected over 6 years using two MRI scanners of different main magnetic fields: 1.5 T Siemens Area and 3.0 T Siemens Trio Tim (Siemens Medical Solutions, Germany)~\cite{bernard2018deep}.  

\vspace{0.2em}\noindent\textbf{Kaggle Data Science Bowl Cardiac Challenge Dataset:}\hspace{0.5em}It includes short-axis view CMR videos from 1100 subjects. Each subject contains multiple slices (8-10) and each slice is a video sequence covering at least one cardiac cycle. Overall, there are 11202 videos in this dataset. The original challenge is to predict the ejection fraction from the videos. Ejection fraction ground truth was provided but irrelevant to our study. The subjects have a large health and age range and the images were collected from numerous sites~\cite{kaggle}. However detailed information such as disease types is not disclosed nor there are segmentation labels. Nevertheless, this large real clinical dataset can be used to train the baseline dense motion model.

\begin{figure*}[ht]
\centering
\subfigure{
\label{frame1} 
\includegraphics[width=.8\linewidth]{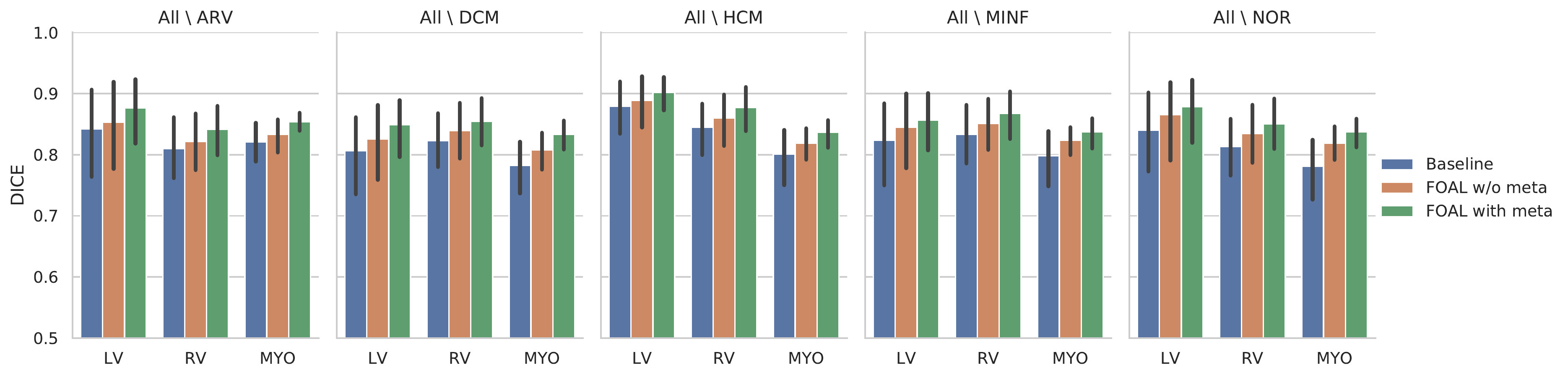}}
\subfigure{
\label{frame2} 
\includegraphics[width=.8\linewidth]{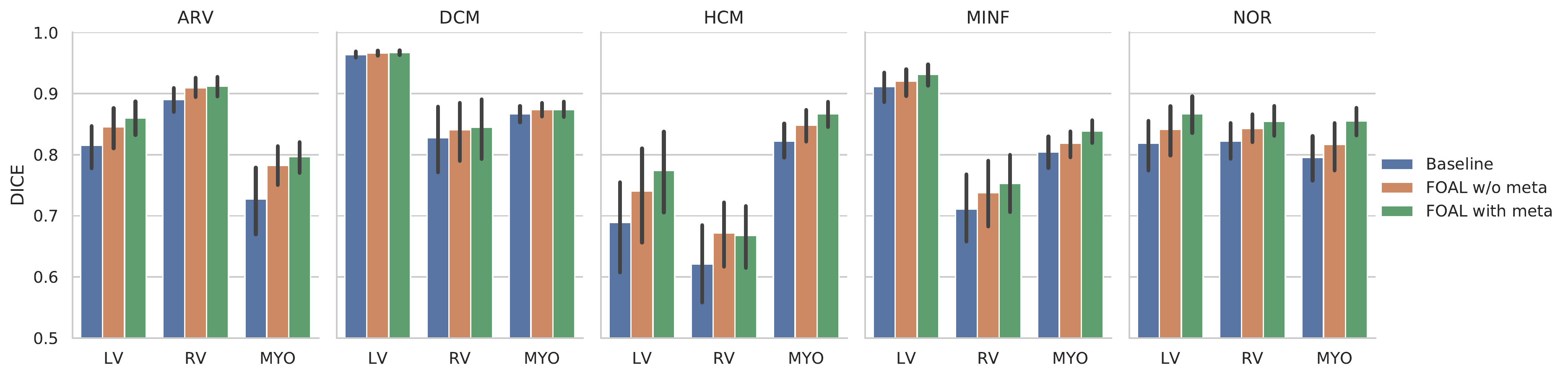}}
\caption{The bar-plots of inside distribution v.s. outside distribution Dice coefficient results for the baseline model, proposed FOAL without meta-learning (FOAL w/o meta) and FOAL with meta-learning (FOAL with meta) for all five folds. Different diseases as outside distributions are presented in different columns. The top row is the inside distribution test and the bottom row is the outside distribution test. The data of the outside distribution disease were excluded in the baseline training and meta-training. Averaged values and standard deviations are presented. }
\label{fig:in_out_acdc}
\end{figure*}

\subsection{Implementation Details}
For image preprocessing, we normalized the gray value to 0-255 and we applied center cropping and zero padding to adjust image size to $192\times{192}$. All models are trained and tested on a Tesla V100 workstation. The other implementation details are presented as following.

\vspace{0.2em}\noindent\textbf{Dense motion tracker:}\hspace{0.5em} As for the baseline model, we adopted a lightweight (shallower and narrower) version of the motion prediction network proposed by Qin~\etal~\cite{qin2018joint}. We halved the number of feature maps of each layer and the number of layers. We set $\alpha_s=5\times\mathrm{10}^{-5}$ and $\beta_c=\mathrm{10}^{-6}~$ in Eq.~\eqref{eq:total_loss}. The batch size is 20 images. We utilized Adam optimizer with an initial learning rate $\mathrm{10}^{-3}$. 

\vspace{0.2em}\noindent\textbf{Online optimizer:}\hspace{0.5em}The number of update steps $m=3$ and the number of sampled pairs $K=24$ in Algorithm~\ref{alg:online}.  We used Adam optimizer with learning rate $\alpha=\mathrm{10}^{-4}$.

\vspace{0.2em}\noindent\textbf{Meta learner:}\hspace{0.5em}We used the number of sampled videos $n=2$, the number of update steps $m=5$, and the number of sampled pairs $K=24$ in the online optimization in Algorithm~\ref{alg:meta-train}. SGD optimizer is used for online optimizer with a fixed learning rate $\alpha=\mathrm{10}^{-5}$. Adam optimizer is used for the meta-learner with an initial learning rate  $\beta=\mathrm{10}^{-5}$ in Algorithm \ref{alg:meta-train}. The meta training steps are 6,000. 

\subsection{Experiment Setups}\label{sec:exp_set}
\vspace{0.2em}\noindent\textbf{Inside distribution vs Outside distribution:}\hspace{0.5em}In data-driven machine learning, we always hypothesize that training samples and testing samples are drawn from the same distribution (inside distribution). The violation of the hypothesis (outside distribution in the testing set) usually gives poor model generalization on the testing set. In this study, we performed five-fold cross-validations in light of the leaving-one-disease-out method on the ACDC dataset. The idea behind this is to separate inside distribution ($P_{in}$) and outside distribution ($P_{out}$) in terms of known diseases. Due to the significant cardiac anatomy and dynamic differences between different diseases, one disease category could be viewed as an outside distribution compared to the other 4 diseases. For subjects in the inside distribution set, we separate them into train set ($80\times80\%=64$ subjects) as $p(D_{a})$ and $p(D_{meta})$, and test set ($80\times20\%=16$ subjects) as $p(D_{t_{inside}})$. $100\%$ subjects in the outside distribution (20 subjects) set were used in the test set as $p(D_{t_{outside}})$. In this experiment, we trained and evaluated all three compared methods on the ACDC dataset.

\vspace{0.2em}\noindent\textbf{Fine-tuning and Generalization:}\hspace{0.5em}We observed that the proposed FOAL with meta-learning needs to train the meta-learner from a baseline model. In the dense tracking context, it is difficult to train the meta-learner from scratch. However, our idea behind the FOAL is to enable any dense tracker to boost their performance via online optimization through meta-learning. To validate the generalizability, we utilized the Kaggle dataset that is without any meta information. Specifically, we used the $30\%$ subjects of the entire Kaggle dataset as $p(D_{a})$ to train the baseline model. We then performed leave-one-disease-out cross-validation on the ACDC dataset. Note that the Kaggle data are only used for training the baseline model while $p(D_{meta})$, $p(D_{t_{inside}})$ and $p(D_{t_{outside}})$ are all from ACDC with the same split in the first experiment. 
In addition to the leave-one-disease-out cross-validation, starting from the baseline model trained on Kaggle, we also compared a vanilla fine-tuning model to FOAL with meta-learning using $20\%$ of the entire ACDC dataset ($100\times20\%=20$ subjects). $100\%$ or $10\%$ of the rest ACDC data were used to train the two models. All 5 categories were mixed. 

The vanilla fine-tuning model used the same training parameters as the baseline model except that we changed the learning rate to $10^{-5}$ to prevent large parameter drift \cite{li2017learning}.

\subsection{Quantitative Metrics}\label{sec:quant_indices}
We used the DICE coefficient (Eq.~\eqref{eq:dice}) and Hausdorff distance error (Eq.~\eqref{eq:Hausdorff}) as quantitative metrics to evaluate the compared tracking methods on segmentation masks. The metrics are defined as: 
\begin{equation}
\label{eq:dice}
    DICE = \frac{ 2 \times |S_{A}\cap{S_{B}}|}{|S_{A}| + |S_{B}|},
\end{equation}
where $S_{A}$ and $S_{B}$ are the segmentation mask A and the segmentation mask B, respectively.
\begin{equation}
\label{eq:Hausdorff}
    H(C_{A}, C_{B}) = \max_{ a \in C_{A}}\{\min_{ b \in C_{B}} ||a-b||_2 \},
\end{equation}
where $a$ and $b$ are the points on the contour A and the contour B, respectively. $||\cdot||_2$ is the Euclidean distance.

\section{Results and Discussion}

\vspace{0.2em}\noindent\textbf{Inside distribution vs outside distribution on ACDC data:}\hspace{0.5em}The five-fold cross-validation experiment in this part is described in Section~\ref{sec:exp_set}. Fig.~\ref{fig:in_out_acdc} depicts all three compared methods (baseline model, FOAL without meta-learning and FOAL with meta-learning) in every cross-validation with test samples drawn from inside or outside distribution. Table~\ref{tab:acdc_dice} and Table~\ref{tab:acdc_hd} summarize Dice and Hausdorff distance results, respectively, for both inside and outside distributions averaged over the five folds. Fig.~\ref{fig:in_out_acdc}, Table~\ref{tab:acdc_dice} and Table~\ref{tab:acdc_hd} show that the proposed FOAL with meta-learning approach outperforms the baseline tracker. For the inside distribution test, our FOAL with meta-learning increased the Dice by $3.7\%$ and reduced Hausdorff distance error by $1.0~mm$ on average. It is worth pointing out that even the training and testing are within the same disease distribution, the variations from patients, scanner types, scanner settings, etc. are still large, which can explain the reduced errors from our method compared to the baseline. The largest accuracy improvement occurs on MYO with $4.3\%$ on Dice for both inside distribution and outside distribution. On the zero-shot (outside distribution) dataset, our FOAL with meta-learning achieves superior performance (e.g. on average $3.8\%$ increase on Dice) compared to the baseline. Besides, we observed that FOAL with meta-learning outperforms FOAL without meta-learning consistently. This demonstrates the effectiveness of meta-learning to enhance the adaptation capability of the online optimizer. This result is not surprising because the online optimizer learns how to adapt to a new video using offline meta training on a large number of videos. This capability teaches the online optimizer to find a sub-optimal path to a better solution than the optimizer without meta-learning can.

\begin{figure*}[ht]
\centering
\subfigure{
\includegraphics[width=.14\linewidth]{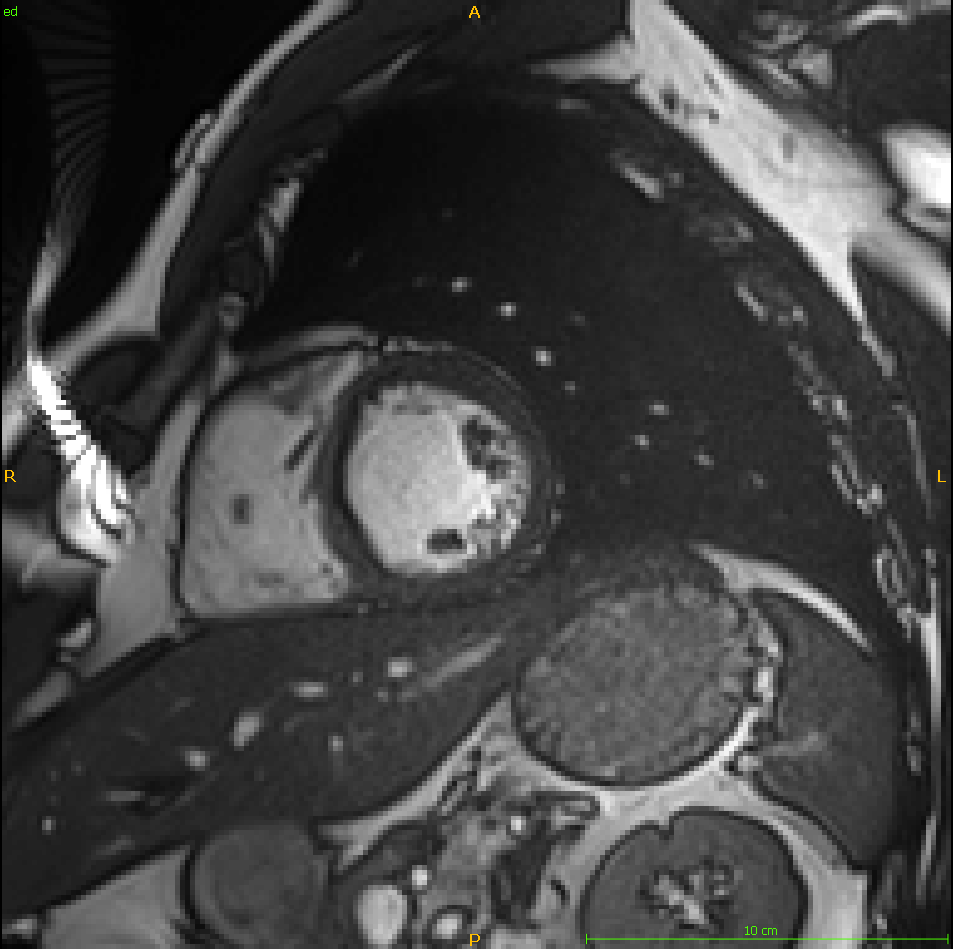}}
\subfigure{
\includegraphics[width=.14\linewidth]{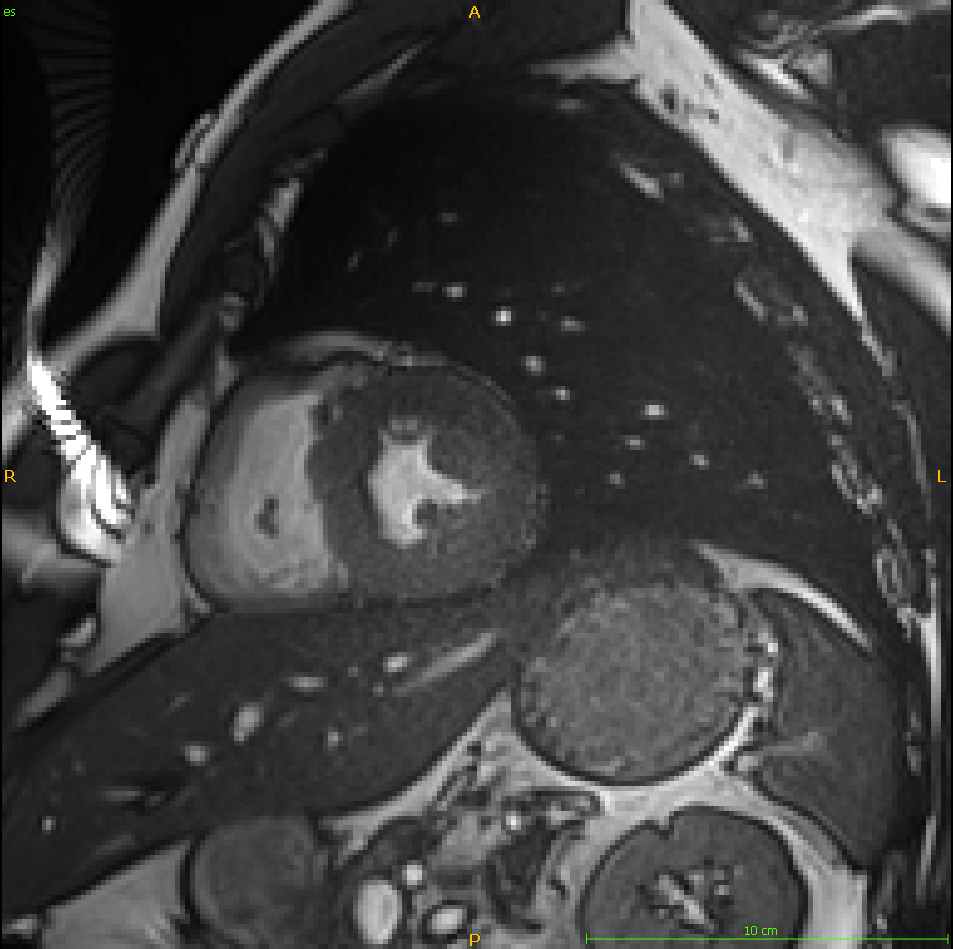}}
\subfigure{
\includegraphics[width=.14\linewidth]{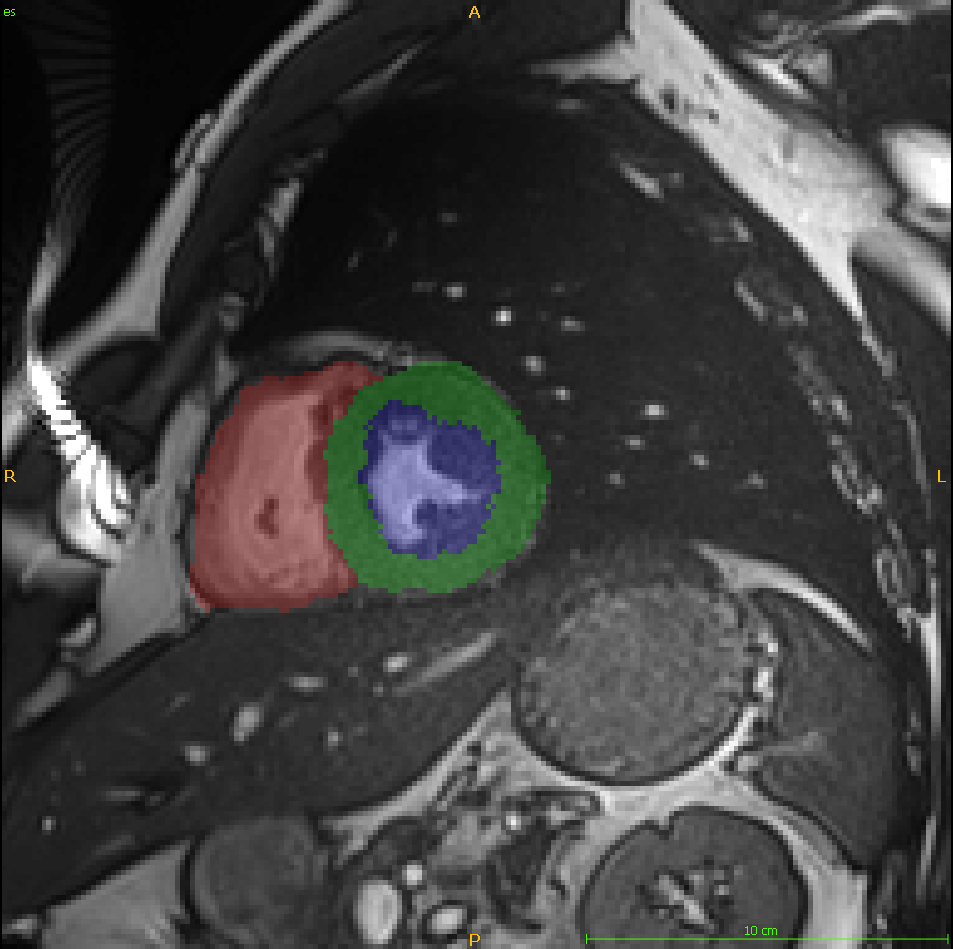}}
\subfigure{
\includegraphics[width=0.14\linewidth]{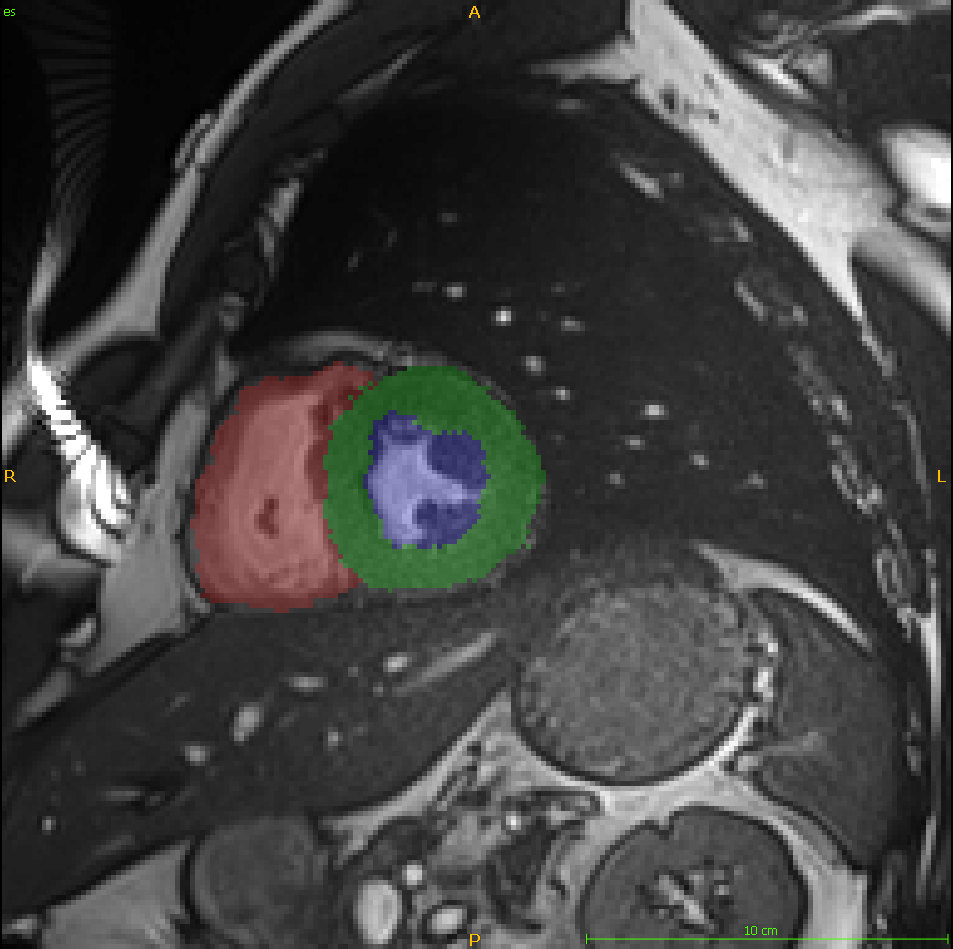}}
\subfigure{
\includegraphics[width=0.14\linewidth]{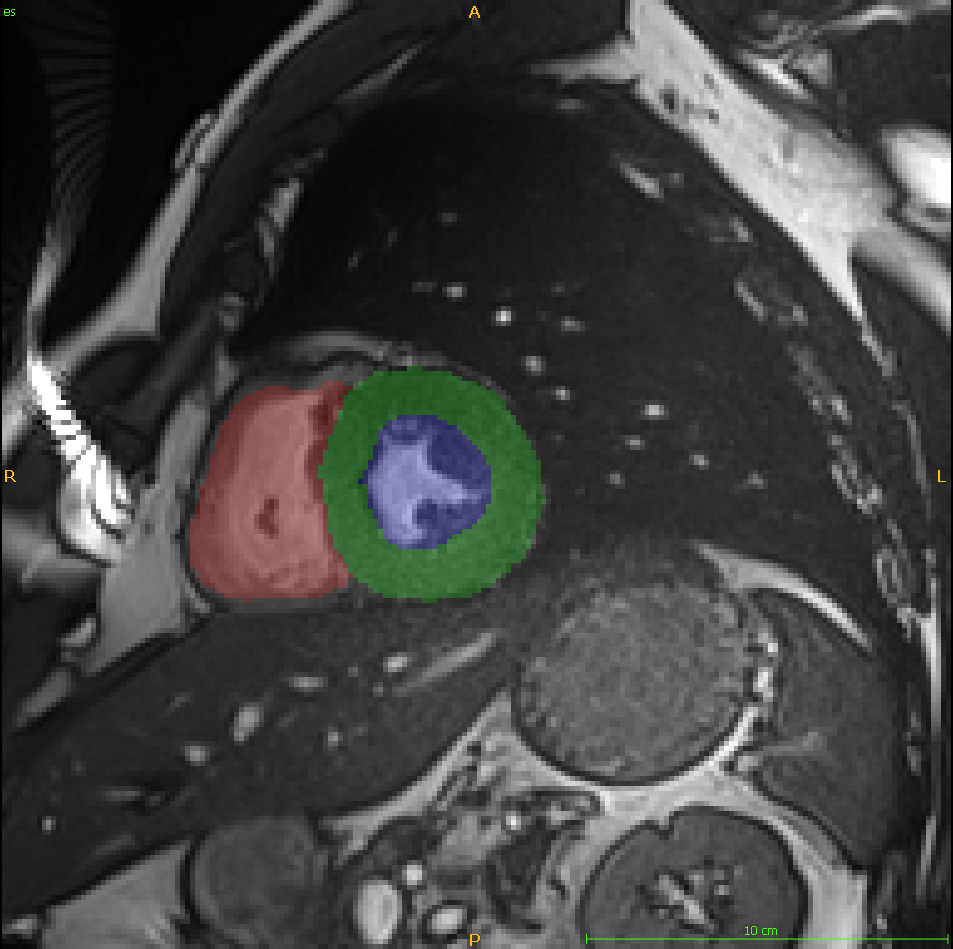}}

\subfigure{
\includegraphics[width=.14\linewidth]{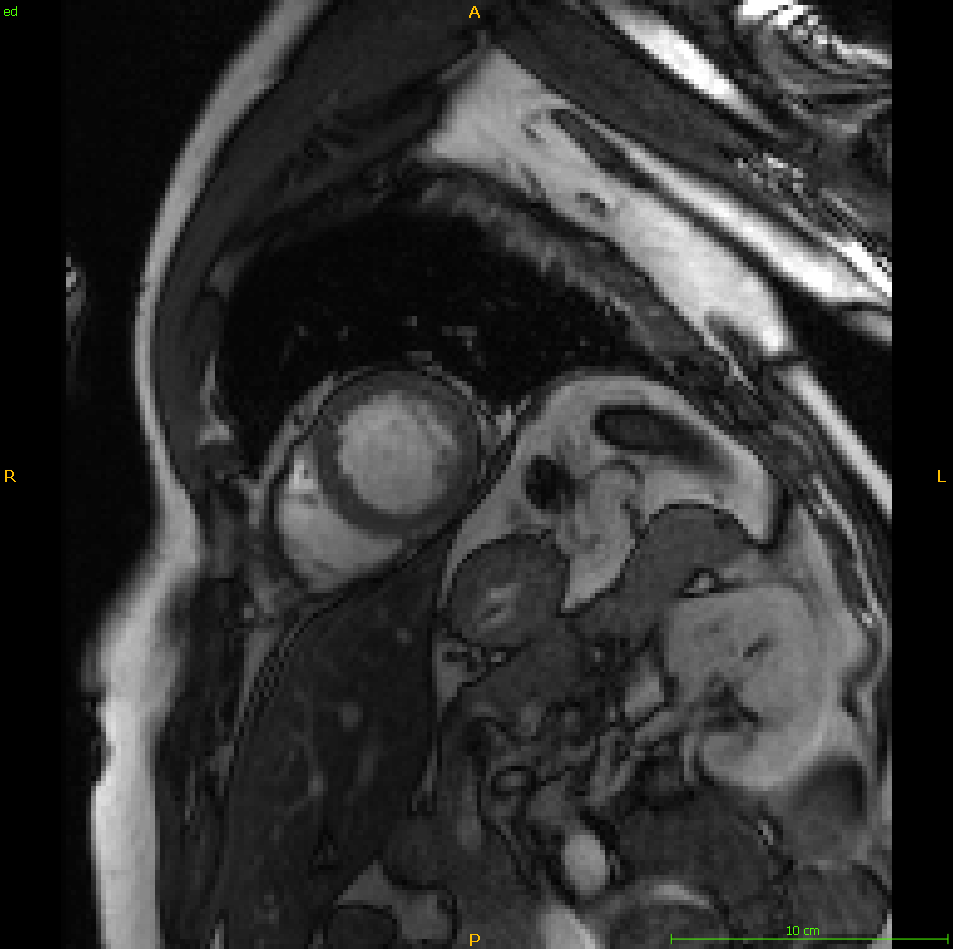}}
\subfigure{
\includegraphics[width=.14\linewidth]{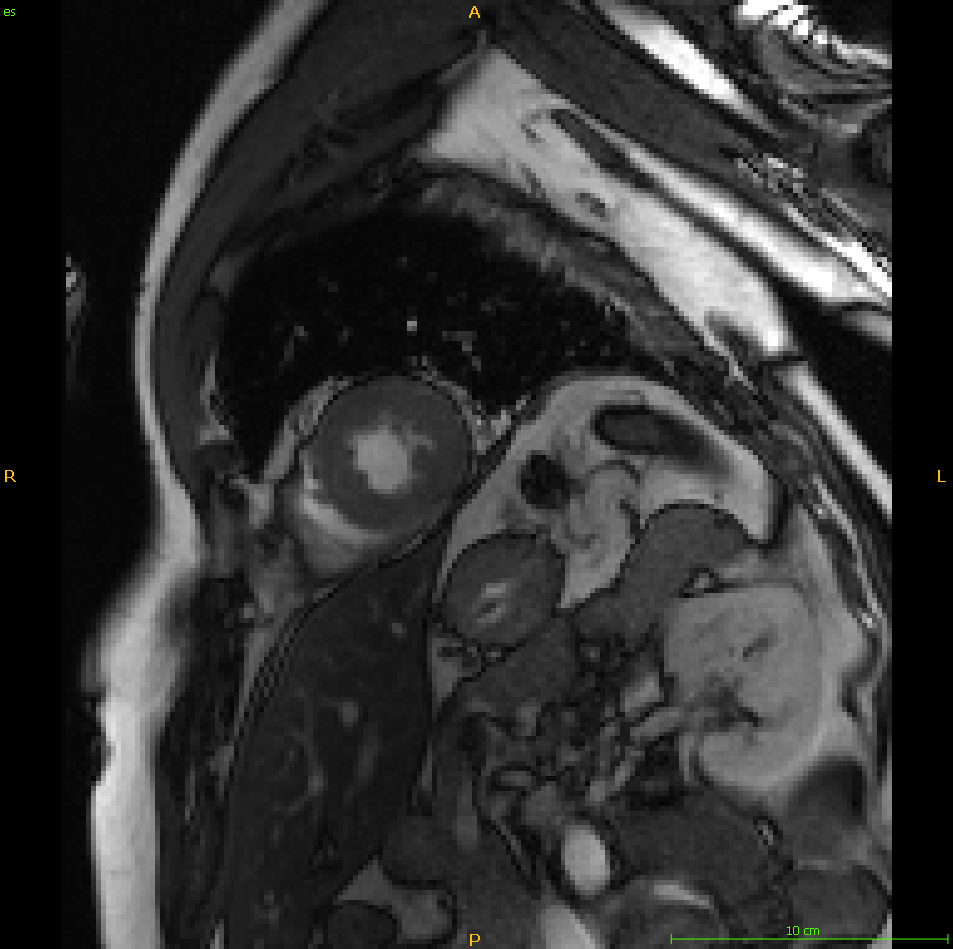}}
\subfigure{
\includegraphics[width=.14\linewidth]{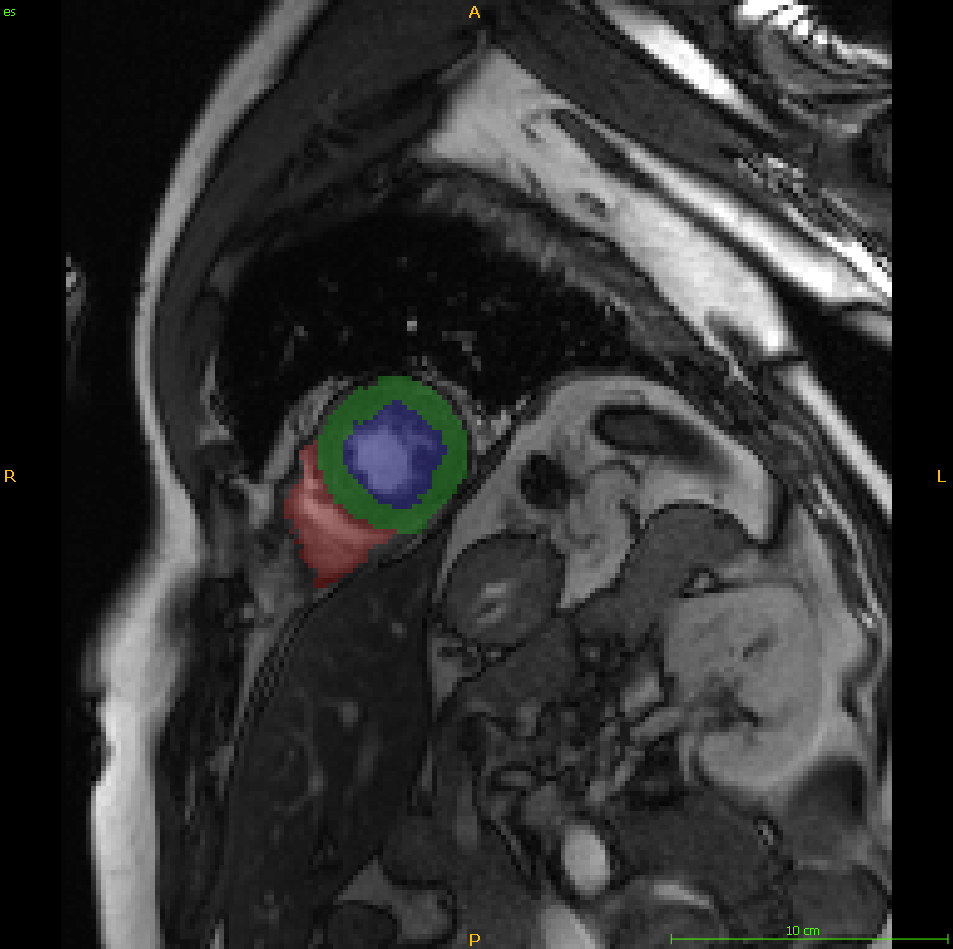}}
\subfigure{
\includegraphics[width=.14\linewidth]{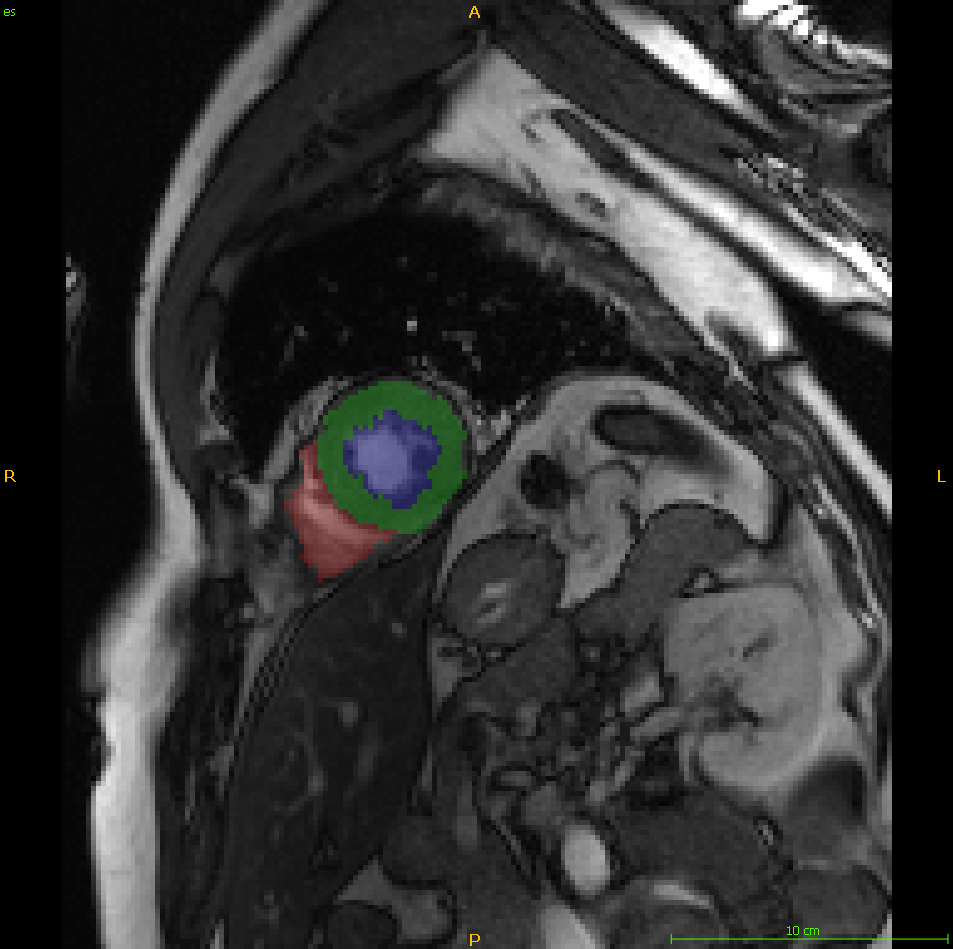}}
\subfigure{
\includegraphics[width=.14\linewidth]{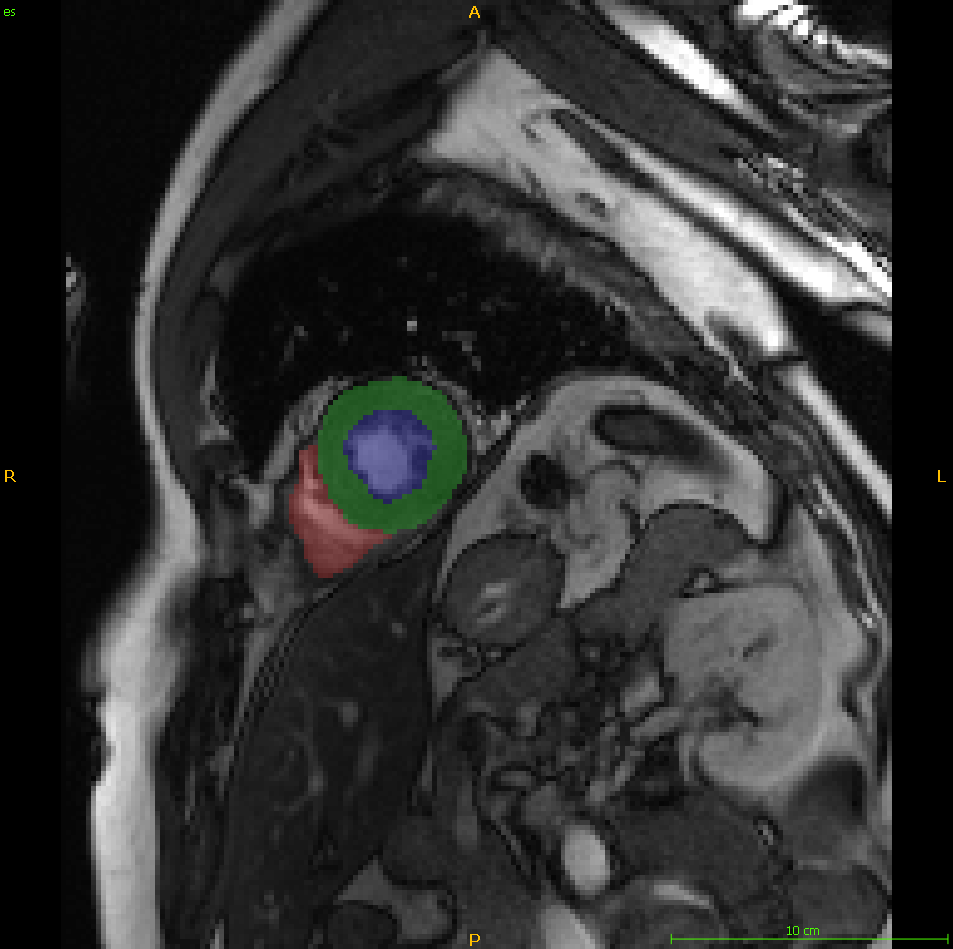}}

\subfigure{
\includegraphics[width=0.14\linewidth]{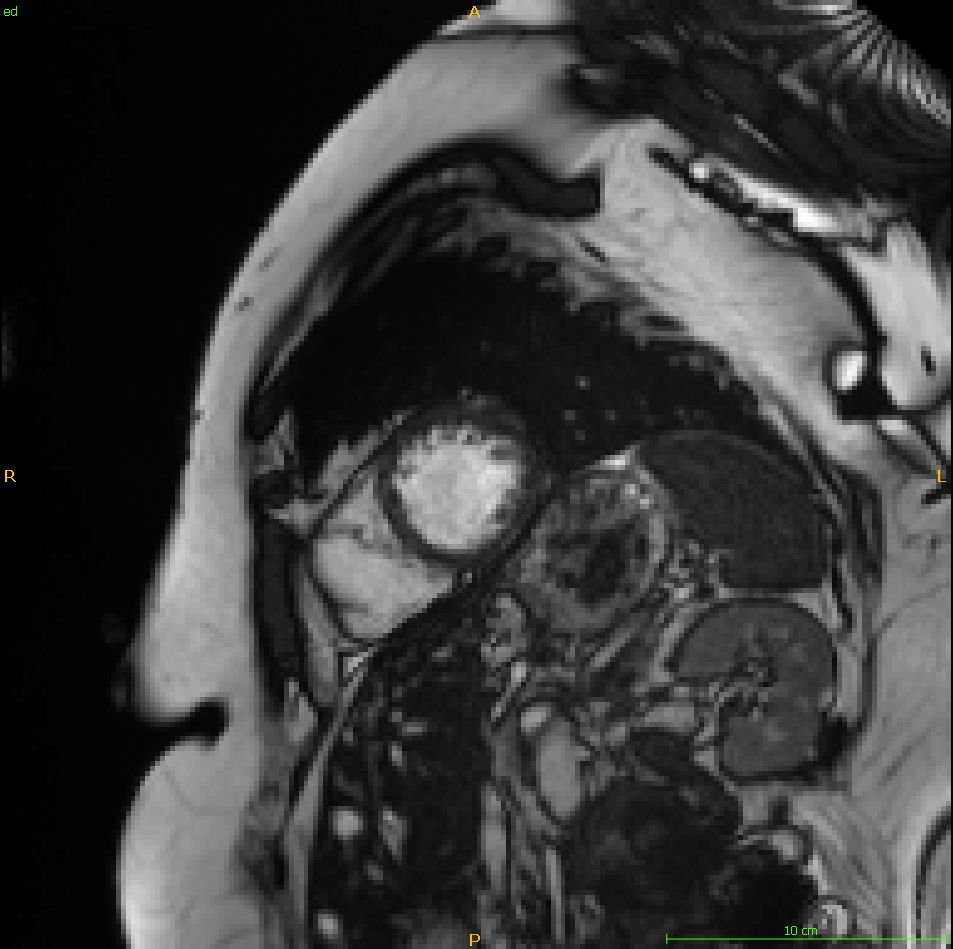}}
\subfigure{
\includegraphics[width=0.14\linewidth]{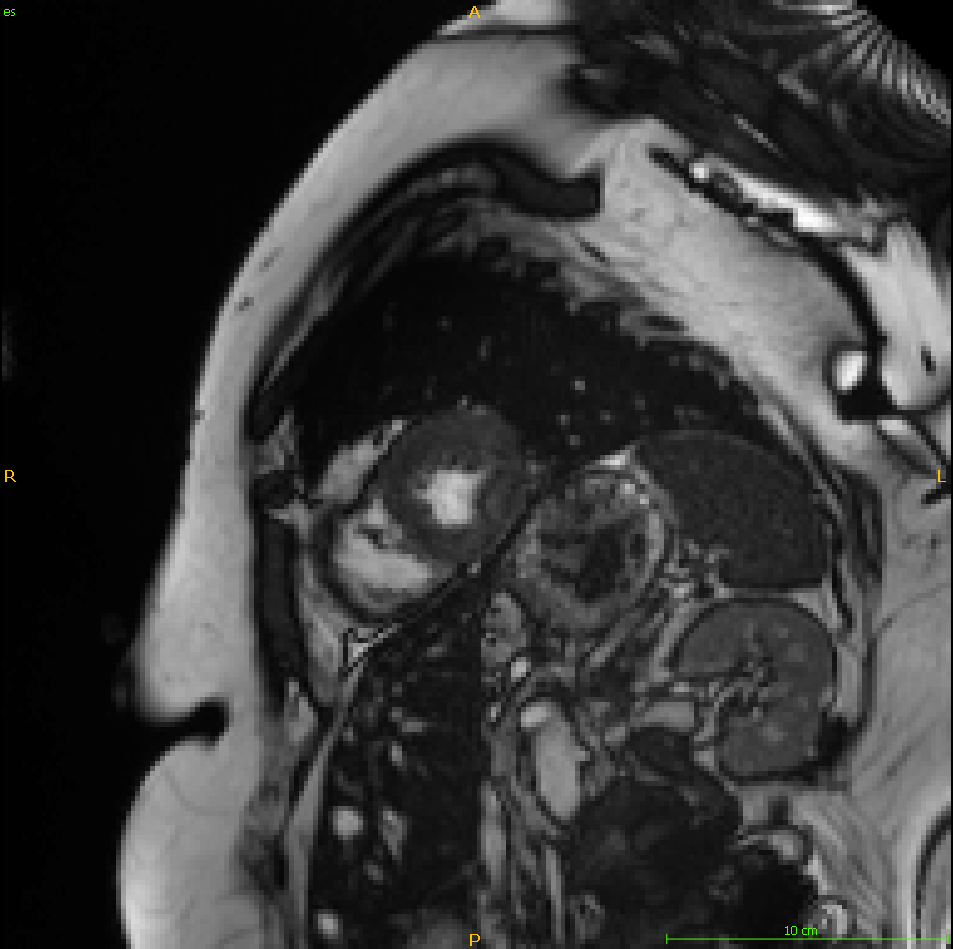}}
\subfigure{
\includegraphics[width=0.14\linewidth]{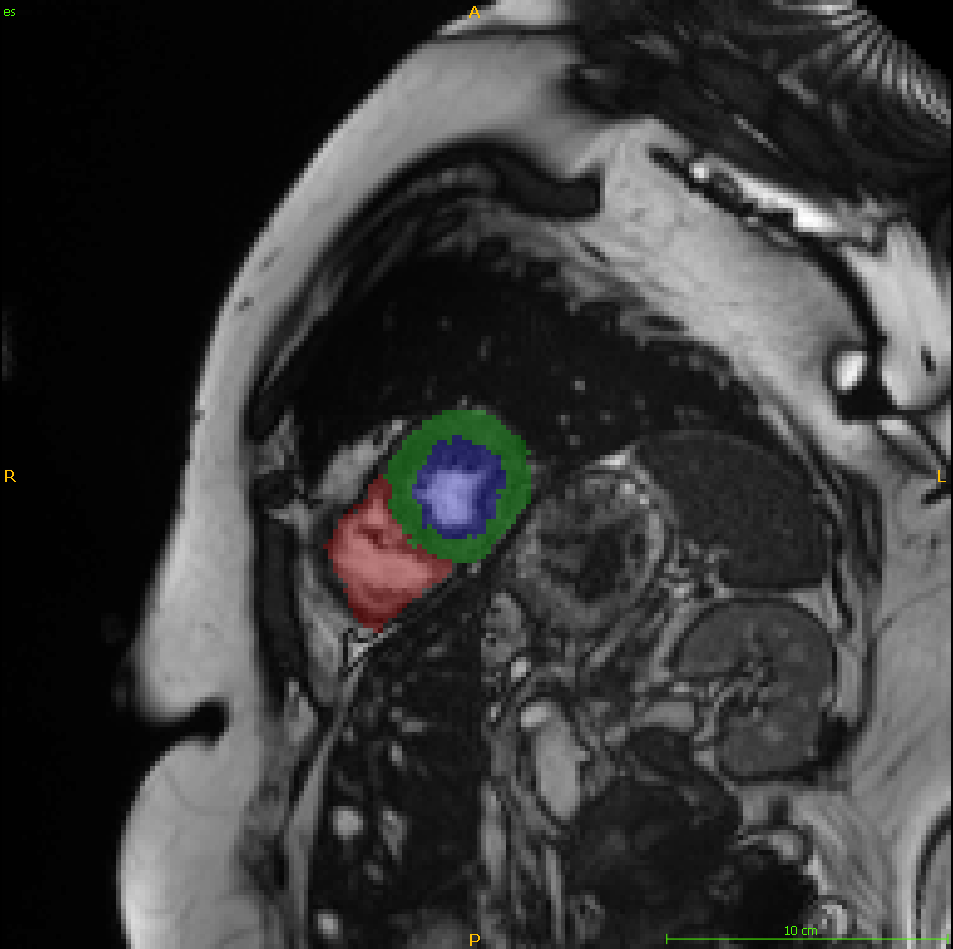}}
\subfigure{
\includegraphics[width=0.14\linewidth]{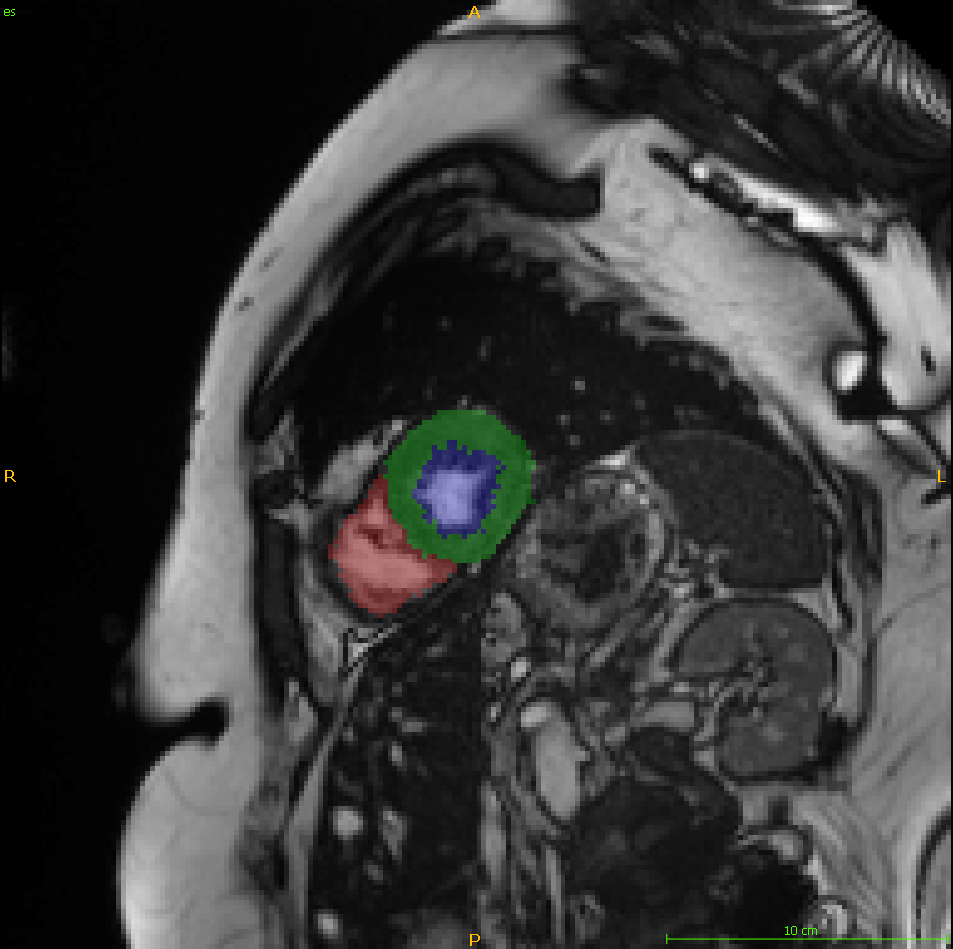}}
\subfigure{
\includegraphics[width=0.14\linewidth]{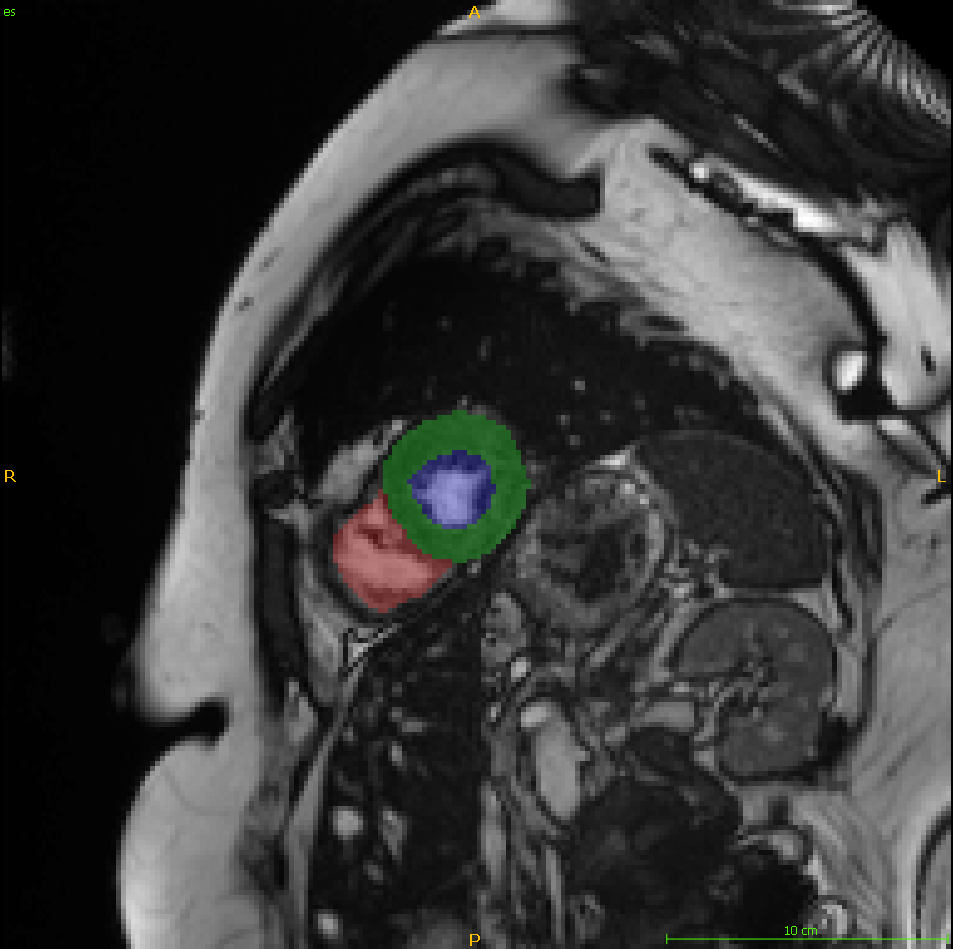}}

\caption{Examples of the tracking results of the mask overlay warped from ED heart phase to ES heart phase. The warp operation utilized deformation fields which were generated from the compared methods. From left to right: the starting frame (ED phase), the final frame (ES phase), baseline model, FOAL with meta-learning and the expert mask annotations. Note that the red mask represents RV, green represents MYO and blue represents LV.}
\label{fig:results_vis} 
\vspace{-1em}
\end{figure*}

Fig.~\ref{fig:results_vis} depicts the warped segmentation results using corresponding deformation fields which were generated by the baseline model and FOAL with meta-learning. In Fig.~\ref{fig:results_vis}, ED and ES frames in the video are also illustrated. We observed a significant appearance and shape difference inside the heart region. Referring to annotations, our method improved LV (blue color) and MYO (green color) comparing to the baseline method. Note that the result can not be compared directly with the results in supervised segmentation \cite{qin2018joint} since our task is unsupervised motion tracking.

\begin{table}[t]
\small
 \caption{Finetuning experiment with Kaggle baseline training and ACDC inside and outside distribution test sets. Dice coefficients are averaged over the five-fold cross-validation for baseline model trained on Kaggle data (Baseline), fine-tuned model on the ACDC dataset (Finetune) and FOAL with meta-learning (FOAL + meta) on the ACDC dataset. Numbers are shown in mean(std).}
 \label{tab:kaggle_dice}
 \centering
 \begin{tabular}{cccc}
  \toprule
  \multirow{2}{*}{Method}   & LV& RV & MYO \\
  \cmidrule(r){2-4} & \multicolumn{3}{c}{Inside Distribution Test Set}\\
  \midrule
 Baseline & 0.864(0.019) & 0.847(0.013) & 0.830(0.010) \\
Finetune &  0.861(0.023) & 0.850(0.012) & 0.827(0.014)\\
 FOAL + meta & \textbf{0.880(0.017)} & \textbf{0.866(0.010)} & \textbf{0.847(0.009)} \\
 \hline
      & \multicolumn{3}{c}{Outside Distribution Test Set}\\
 \hline 
Baseline & 0.874(0.070) & 0.796(0.093) & 0.841(0.024) \\
 Finetune & 0.870(0.070) & 0.792(0.094) & 0.833(0.031)\\
 FOAL + meta & \textbf{0.885(0.059)} & \textbf{0.804(0.091)} & \textbf{0.849(0.023)} \\      
  \bottomrule
 \end{tabular}
\end{table}

\begin{table}[ht]
\small

 \caption{Finetuning experiment with Kaggle baseline training and $100\%$ and $10\%$ ACDC training dataset. Dice coefficients (mean(std)) for baseline model trained on Kaggle data (Baseline), vanilla fine-tuned model on the ACDC (Finetune) and FOAL with meta-learning on the ACDC (FOAL + meta) are reported.}
 \label{tab:kaggle_10_100}
 \centering
 \begin{tabular}{cccc}
  \toprule
  \multirow{2}{*}{Method}   & LV& RV & MYO \\
  \cmidrule(r){2-4} & \multicolumn{3}{c}{$100\%$ of ACDC training data}\\
  \midrule
 Baseline & 0.865(0.103) & 0.845(0.080) & 0.829(0.065) \\
 Finetune & 0.865(0.104) & 0.854(0.079) & 0.831(0.063)\\
 FOAL & \textbf{0.881(0.086)} & \textbf{0.865(0.070)} & \textbf{0.845(0.051)}\\
 \hline
      & \multicolumn{3}{c}{$10\%$ of ACDC training data}\\
 \hline 
 Baseline & 0.865(0.103) & 0.845(0.080) & 0.829(0.065) \\
 Finetune & 0.864(0.104) & 0.845(0.082) & 0.824(0.073)\\
 FOAL +meta & \textbf{0.882(0.086)} & \textbf{0.863(0.071)} & \textbf{0.845(0.051)}\\
  \bottomrule
 \end{tabular}
\end{table}

\vspace{0.2em}\noindent\textbf{Fine-tuning and Generalization:}\hspace{0.5em}The experiment setup in this part is discussed in Section~\ref{sec:exp_set}. We compared the baseline model trained on Kaggle data (Baseline), a model fine-tuned on ACDC data from the baseline model (Finetune) and our proposed FOAL with meta-learning from the baseline model (FOAL+meta). Averaged Dice coefficients among five folds for both inside distribution and outside distribution can be found in Table~\ref{tab:kaggle_dice}. The baseline model performs comparably well on both distributions except RV. This might be because the Kaggle dataset consists of a variety of cardiac diseases and it has distribution overlaps with both the inside distribution and the outside distribution datasets but not for RV. Fine-tuning the model on the ACDC dataset does not improve the performance. Comparing to the baseline model, our method improved $2.7\%$ on the inside distribution test and $2.4\%$ on the outside distribution test in terms of Dice. Though we test the generalization of the method on CMR datasets, FOAL may have the potential  of generalization to other motion(flow) estimation datasets like the KITTI and Sintel Final.

Table~\ref{tab:kaggle_10_100} shows Dice results for vanilla fine-tuning model and our FOAL with meta-learning using $10\%$ or $100\%$ ACDC training samples. In contrast to the leave-one-disease-out experiments, we did not isolate any disease in the training samples in this experiment and the models were tested on the entire ACDC test set. Vanilla fine-tuning model made the performance slightly worse in the $10\%$ experiment while it slightly improved the accuracy in the $100\%$ experiment comparing to the baseline model. Meanwhile, FOAL with meta-learning gave $1.68\%$ and $1.71\%$ Dice increases on average for both $10\%$ and $100\%$ experiments, respectively. This result is consistent with the above fivefold cross-validation test. In addition, Fig.~\ref{tab:kaggle_10_100} demonstrates that our FOAL performs comparably well using a small amount of data when it is meta-trained from a strong baseline model.

Our FOAL online optimization algorithm requires $413\pm{8}$ milliseconds (mean$\pm$standard deviation), which we find it completely durable for most current clinical applications. 

\vspace{-1mm}
\section{Conclusion}
\vspace{-1mm}
In this work, we proposed a novel online adaptive learning method to minimize the domain mismatch problem in the context of dense cardiac motion estimation. The online adaptor is a gradient descent based optimizer which itself is also optimized by a meta-learner. The meta-learning strategy allows the online optimizer to perform a fast adaption using a limited number of model updates and a small number of image pairs from a single video. The tracking performance is significantly improved in all the zero-shot (outside distribution comparing to the training samples) experimental setups. Also, it is observed that the online adaptor can minimize the tracking errors in the inside distribution tests. Experimental results demonstrate that our methods obtain superior performance compared to the model without online adaption. The pilot study shows the feasibility of applying the method in the context of unsupervised dense motion tracking or deformable image registration. The proposed method provides a practical and elegant approach to an often overlooked problem in existing art. We hope to inspire more discussions and work to benefit other clinical applications suffering from similar issues.

\clearpage
{\small
\bibliographystyle{ieee_fullname}
\bibliography{egbib}
}

\end{document}